\documentclass[11pt]{article}

\usepackage[preprint,nohyperref]{acl}
\usepackage{blindtext}
\usepackage{times}
\usepackage{latexsym}
\usepackage{stfloats}

\usepackage[T1]{fontenc}

\usepackage[utf8]{inputenc}

\usepackage{microtype}

\usepackage{inconsolata}

\usepackage{graphicx}
\usepackage{epstopdf} 
\usepackage{comment}
\usepackage{booktabs}  
\usepackage{caption}   
\usepackage{tabularx}

\usepackage{phew}
\usepackage{mdframed}

\usepackage{tabularx}
\usepackage{graphicx}
\usepackage{rotating}

\newcolumntype{L}{>{\raggedright\arraybackslash}X} 

\title{Supervised Fine-Tuning versus Reinforcement Learning: \\ A Study of Post-Training Methods for Large Language Models}

\author{\textbf{Haitao Jiang$^{1}$, \quad Wenbo Zhang$^{2}$, \quad Jiarui Yao$^{3}$, \quad Hengrui Cai$^{2}$, \quad Sheng Wang$^{4}$, \quad Rui Song$^{5}$}\\[0.3em]
$^{1}$ North Carolina State University \quad $^{2}$ UC Irvine \quad $^{3}$ University of Illinois Urbana-Champaign \\ 
$^{4}$ University of Washington \quad $^{5}$ Amazon \\
}

\usepackage[acronym]{glossaries}
\makeglossaries

\setacronymstyle{long-short}

\usepackage[colorlinks]{hyperref}
\hypersetup{
    linkcolor=black,    
    citecolor=blue,
    urlcolor=blue
}

\newacronym{llm}{LLM}{Large Language Model}
\newacronym{sft}{SFT}{Supervised Fine-Tuning}
\newacronym{rl}{RL}{Reinforcement Learning}
\newacronym{cot}{CoT}{Chain-of-Thought}
\newacronym{bc}{BC}{Behavior Cloning}
\newacronym{gem}{GEM}{Entropic Distribution Matching}
\newacronym{ppo}{PPO}{Proximal Policy Optimization}
\newacronym{grpo}{GRPO}{Group Relative Policy Optimization}
\newacronym{kl}{KL}{Kullback–Leibler}
\newacronym{rlhf}{RLHF}{Reinforcement Learning from Human Feedback}
\newacronym{vlm}{VLMs}{Vision Language Models}
\newacronym{dpo}{DPO}{Direct Preference Optimization}
\newacronym{qa}{QA}{Question Answering}
\newacronym{rag}{RAG}{Retrieval-Augmented Generation}
\newacronym{fm}{FM}{Foundation Model}
\newacronym{rft}{RFT}{Rejection Fine-Tuning}

\begin{document}
\maketitle

\begin{abstract}
Pre-trained \gls{llm} exhibits broad capabilities, yet, for specific tasks or domains their attainment of higher accuracy and more reliable reasoning generally depends on post-training through \gls{sft} or \gls{rl}.
Although often treated as distinct methodologies, recent theoretical and empirical developments demonstrate that \gls{sft} and \gls{rl} are closely connected. 
This study presents a comprehensive and unified perspective on \gls{llm} post-training with \gls{sft} and \gls{rl}. We first provide an in-depth overview of both techniques, examining their objectives, algorithmic structures, and data requirements. We then systematically analyze their interplay, highlighting frameworks that integrate \gls{sft} and \gls{rl}, hybrid training pipelines, and methods that leverage their complementary strengths. Drawing on a representative set of recent application studies from 2023 to 2025, we identify emerging trends, characterize the rapid shift toward hybrid post-training paradigms, and distill key takeaways that clarify when and why each method is most effective.
By synthesizing theoretical insights, practical methodologies, and empirical evidence, this study establishes a coherent understanding of \gls{sft} and \gls{rl} within a unified framework and outlines promising directions for future research in scalable, efficient, and generalizable \gls{llm} post-training.
\end{abstract}

\begin{figure*}[t]
\centering
\begin{tikzpicture}[
    scale=0.85,
    transform shape,
    node distance=1.5cm and 2.5cm,
    box/.style={rectangle, draw=#1, rounded corners, minimum width=3.5cm, minimum height=0.8cm, align=center, font=\footnotesize}, 
    arrow/.style={->, thick, >=stealth}
]

\node (m1) [box=black, anchor=west, text width=15cm, align=left] at (0,0) {\textbf{SFT.} Algorithm-centric: \citet{Li2024EntropicDistributionMatching}, \citet{Pang2025TokenCleaning}, \citet{Ming2025OneTokenRollout} ; Data-centric: \citet{zhou2023lima}, \citet{deb2025fishersft}, \citet{li2025data}, \citet{quan2025automatically}, \citet{cao2025condor}.
};
\node (m2) [
    box=black, 
    below=1.5cm of m1.west, 
    anchor=west,
    text width=15cm,   
    align=left         
] {\textbf{RL.} Algorithm-centric: 
\citet{li2023remax}, \citet{shao2024deepseekmath}, \citet{ahmadian2024back}, \citet{hu2025reinforce++}, \citet{cheng2025reasoning}, \citet{cui2025entropy}, \citet{he2025skywork}, \citet{shrivastava2025sample}, \citet{chen2025seed} ; Data-centric: \citet{zhang2024policy}, \citet{xu2025not}, \citet{zheng2025act}, \citet{qu2025can}, \citet{zhang2025clpo}, \citet{chen2025self}, \citet{wang2025dump}.
};


\node (m3) [
    box=black, 
    below=1.7cm of m2.west, 
    anchor=west,
    text width=15cm,   
    align=left         
] {
\textbf{Unifying and Cross-Enhancing SFT \& RL.} 
    {A Unified Objective:}  \citet{wu2025generalization}, \citet{lv2025towards}, \citet{liu2025uft};
    {Using SFT to Enhance RL:} \citet{li2024getting}, \citet{hua2024intuitive}, \citet{chen2025bridging};
    {Using RL to Enhance SFT:}  
    \citet{wang2024uft}, \citet{wu2025generalization}, \citet{qin2025supervised}, \citet{du2025simplify}.
};

\node (m5) [box=black, below=1.5cm of m3.west, anchor=west, text width=15cm, align=left] {\textbf{Combining SFT and RL.} Hybrid training with combination of SFT and RL objectives: 
\citet{hong2024q}, \citet{liu2024provably}, \citet{huang2025blending}, \citet{zhang2025policy}, \citet{liu2025uft}, \citet{lv2025towards}, \citet{fu2025srft},  \citet{chen2025step}, \citet{deng2025supervised}};

\node (m6) [
    box=black, 
    below=1.5cm of m5.west, 
    anchor=west,
    text width=15cm,   
    align=left         
] {
    \textbf{General QA Tasks.} 
    CoT Reasoning: \citet{qa_4}, \citet{qa_9}, \citet{qa_11}; 
    LLM-based Retrieval: \citet{qa_10}, \citet{qa_5}, \citet{qa_13}; 
    Hallucination Management: \citet{qa_2}, \citet{qa_6}, \citet{qa_12}.
};

\node (m7) [
    box=black, 
    below=1.5cm of m6.west, 
    anchor=west,
    text width=15cm, 
    align=left
] {
    \textbf{Mathematical Tasks.} 
    Math CoT: \citet{math_data_3}, \citet{math_evol_data_1}, \citet{math_evol_data_2}, \citet{math_data_2}, \citet{shao2024deepseekmath}; 
    Rollout Selection and Verification: \citet{math_verify}, \citet{math_verify_4}, \citet{math_verify_2}, \citet{math_verify_6}, \citet{math_verify_3}.
};

\node (m8) [
    box=black, 
    below=1.5cm of m7.west, 
    anchor=west,
    text width=15cm, 
    align=left
] {
    \textbf{Agentic Tasks.} 
    Task Execution: \citet{rl_data_0}, \citet{sft_data_1}, \citet{sft_data_3}; 
    Self-Improvement: \citet{rl_alg_1}, \citet{both_seq_2},
    \citet{both_sametime_0}, \citet{rl_alg_0}; 
    Planning: \citet{sft_mcts_3}, \citet{both_seq_3}, \citet{sft_mcts_1}.
};

\node (m9) [
    box=black, 
    below=1.5cm of m8.west, 
    anchor=west,
    text width=15cm, 
    align=left
] {
    \textbf{Code-based Tasks.} 
    Code Generation: \citet{code_gen_3}, \citet{code_gen_2}, \citet{code_gen_4}; 
    Code Editing: \citet{code_edit_1}, \citet{code_edit_3}, \citet{code_edit_6}, \citet{code_edit_2}; 
    Code for other tasks: \citet{code_other_1}, \citet{code_other_2}, \citet{code_other_4}.
};


\node (t1) [
    box=black,
    left=1cm of m2,
    yshift=1.8cm,
    rotate=90,
    text width=1.5cm,
    minimum width=2.5cm,
    align=center
] {Overview of 
SFT/RL};

\node (t2) [
    box=black,
    left=1cm of m3,
    minimum width=2.5cm,
    yshift=0.5cm,           
    rotate=90,
    text width=2cm,
    align=center
] {Comparison or
Combination
};

\node (t3) [
    box=black,
    left=1cm of m7,
    minimum width=2.5cm,
    yshift=0.66cm,          
    rotate=90,
    text width=2cm,
    align=center
] {Applications of SFT \& RL};

\node (foundationframe) [
    box=black,
    left=2cm of t2,
    yshift=5cm,           
    rotate=90,
    minimum width=9cm,
    align=center
] {\large Post-training Alignment via SFT and/or RL
};

\draw[arrow=green!60!black] (foundationframe.south) -- (t1.north);
\draw[arrow=blue] (foundationframe.south) -- (t2.north);
\draw[arrow=green!60!black] (foundationframe.south) -- (t3.north);

\draw[arrow=green!60!black] (t1.south) -- (m1.west);
\draw[arrow=blue] (t1.south) -- (m2.west);

\draw[arrow=green!60!black] (t2.south) -- (m3.west);
\draw[arrow=blue] (t2.south) -- (m5.west);

\draw[arrow=green!60!black] (t3.south) -- (m6.west);
\draw[arrow=blue] (t3.south) -- (m7.west);
\draw[arrow=blue] (t3.south) -- (m8.west);
\draw[arrow=blue] (t3.south) -- (m9.west);

\end{tikzpicture}
\caption{A taxonomy of large language model (LLM) post-training alignment methods via supervised fine-tuning (SFT) and reinforcement learning (RL). We organize prior work along three categories: (1) algorithm-centric versus data-centric approaches within SFT and RL, (2) comparative, unifying, and hybrid frameworks that integrate SFT and RL objectives, and (3) representative downstream application domains, including reasoning, mathematics, agentic behavior, and code-related tasks.}
\label{fig:taxonomy}
\end{figure*}
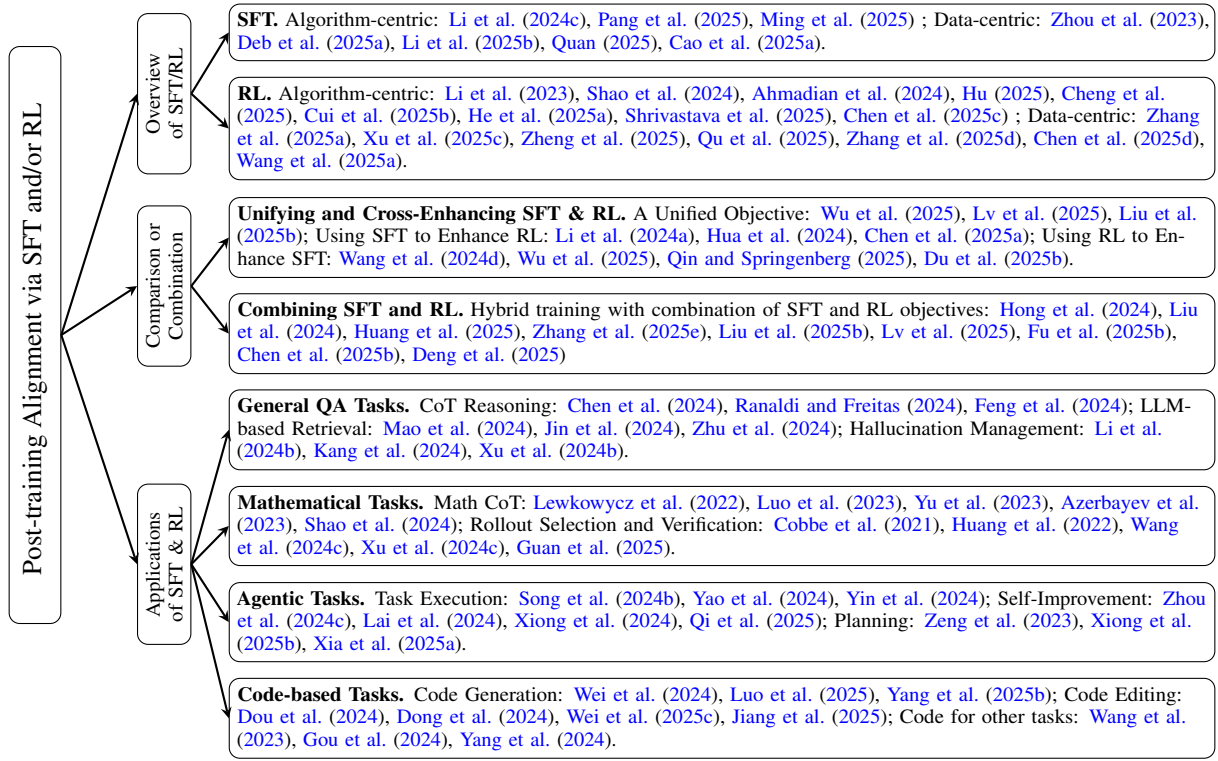

\section{Introduction}
Pre-trained \gls{llm}s have demonstrated remarkable capabilities across a wide range of tasks, from fact-based question answering \citep{joshi2017triviaqalargescaledistantly} to code generation \citep{jimenez2024swebenchlanguagemodelsresolve}.
Despite being trained on corpora containing billions to trillions of tokens, \gls{llm}s often require task-specific post-training adaptation to improve accuracy, mitigate erroneous outputs, and handle new tasks. 
For instance, fine-tuning can enhance multi-step reasoning by enabling the generation of progressively longer reasoning chains, ultimately leading to more accurate final answers, particularly in tasks that require complex reasoning ability\citep{deepseek2024r1}. It can also enhance practical interaction skills, including performing tasks in household \citep{sft_data_6} or device-control environments \citep{rawles2025androidworlddynamicbenchmarkingenvironment}.
Such capabilities rarely emerge from pre-training alone, as the data seldom contain the task-specific patterns or feedback necessary for accurate reasoning or complex interactions, highlighting the importance of post-training.

Current post-training methodologies for \gls{llm}s primarily fall into two paradigms: \gls{sft} and \gls{rl}. 
The objective of \gls{sft} \citep{zhou2023lima, Li2024EntropicDistributionMatching, Pang2025TokenCleaning} is to maximize the likelihood of tokens conditioned on context, whereas \gls{rl} \citep{ouyang2022training,deepseek2024r1,yang2025qwen3} optimizes a reward signal derived from human or automated preference feedback. Despite their different objectives, in recent years, research efforts have increasingly focused on bridging these two approaches \citep{wu2025generalization, fu2025srft}, exploring how their combination \citep{wu2025generalization, qin2025supervised, yan2025learning, liu2025empowering} can enhance performance beyond what either approach achieves in isolation.

This combination is especially pronounced in tasks that demand both accuracy and generalization such as reasoning. \gls{sft} alone can teach models to generate basic \gls{cot}, but may struggle with novel problem structures \citep{ross2010efficient,de2019causal}. Conversely, \gls{rl} fine-tuning based on preference feedback can improve step-wise correctness, yet it often requires extensive exploration in the absence of offline demonstrations. By combining \gls{sft} and \gls{rl}, models can leverage the strengths of both approaches, more reliable and robust reasoning.

As illustrated in Figure \ref{fig:taxonomy}, these studies highlight that understanding and combining the complementary strengths of \gls{sft} and \gls{rl} is crucial for advancing \gls{llm} post-training methods.

While recent studies offer valuable insights into \gls{llm} post-training, the majority of them typically examine \gls{sft} or \gls{rl} separately \citep{parthasarathy2024ultimate, tao2024survey, mao2025survey, tie2025survey, zhang2025survey, zhang-etal-2025-survey-foundation}, leaving the relationships between these approaches comparatively underexplored. Other works focus on specialized dimensions of post-training, such as vision-centric adaptation \citep{chu2025sft}, advances in reasoning \citep{kumar2025llm}, agentic behaviors \citep{du2025survey}, or scaling strategies \citep{lai2025survey}. In contrast, our survey provides a systematic and integrated perspective on \gls{sft} and \gls{rl} as complementary post-training tools, with particular emphasis on their interplay and practical applications.

Our key \textbf{contributions} are threefold:
\begin{itemize}
    \item We are the first study to systematically summarize and compare \gls{sft} and \gls{rl} in \gls{llm} post-training, providing a clear understanding of what \gls{sft} and \gls{rl} are, and how they can be extended from both algorithm-centric and data-centric perspectives.
    \item We then establish a unified framework for characterizing \gls{sft} and \gls{rl}, highlighting how they can complement each other or be integrated into hybrid learning approaches.
    \item From an analysis of applications spanning 2023 to 2025, we observe rapid task-domain expansion, growing adoption of integrated \gls{sft}–\gls{rl} training, and a continued shift from API-based labeling to open-weight–generated datasets.

\end{itemize}

\section{Background: SFT and RL}

\noindent \textbf{Supervised Fine-Tuning (\gls{sft}).} 
\gls{sft} is a method for adapting \gls{llm}s to specific tasks or domains by training on high-quality prompt–response pairs using standard language modeling objectives. This process typically involves collecting or curating a dataset of expert-written demonstrations $D=\{(x,y)\}$, prompt $x$ paired with target responses $y$, and then fine-tuning the model $\pi_\theta$:
$$
\min _\theta \mathbb{E}_{(x, y) \sim \mathcal{D}}\left[-\log \pi_\theta(y \mid x)\right].
$$
In essence, \gls{sft} trains a model to imitate expert behavior. It is closely related to \gls{bc} \citep{pomerleau1991efficient} in traditional reinforcement learning, as both frameworks learn directly from expert demonstrations without relying on explicit reward signals. The goal of \gls{sft} is to reproduce expert performance; however, like \gls{bc}, it suffers from distribution shift, which can lead to compounding errors \citep{ross2010efficient,de2019causal}, and depends heavily on the quality of the demonstrations.

\noindent \textbf{Reinforcement Learning (\gls{rl}).} 
\gls{rl} offers an alternative paradigm in which \gls{llm}s are tuned not through direct supervised signals, but by optimizing behaviors according to a reward function. In classical \gls{rl}, models interact with environments and learn a policy that maximizes cumulative rewards over time. Recent \gls{llm}s post-training methods \citep{ouyang2022training,deepseek2024r1,yang2025qwen3} have adopted similar techniques: an environment is constructed, the model serves as a policy that interacts with this environment, and training proceeds to maximize expected rewards:
\begin{equation*}
    \underset{\theta}{\max} \;
\mathbb{E}_{x \sim \mathcal{D},\, y \sim \pi_\theta(\cdot \mid x)}
\left[ r(x, y) \right],
\end{equation*}
%
where the reward function $r$ is either manually specified or learned from data. Although \gls{rl} can also refer to offline reinforcement learning without environment interactions, in this work, we use it exclusively to denote online \gls{rl}, as it plays a predominant role in the current frontier of \gls{llm} alignment, including reasoning and agent development.


Here, we summarize the key distinction between \gls{sft} and \gls{rl} in the context of \gls{llm} post-training, as commonly described in current literature \citep{shao2024deepseekmath,zhang2025survey}: \gls{sft} is a \textbf{supervised learning} paradigm that trains on expert-annotated prompt–response pairs, Whereas \gls{rl} is a \textbf{reward-driven optimization} paradigm that learns by updating the model from its own generations.





\section{SFT and RL: Distinct Methodological Landscapes}
In the evolving landscape of \gls{llm} post-training, \gls{sft} and \gls{rl} are two primary methodological paradigms. This section presents an overview of these approaches from (1) \textbf{Algorithm-centric}: refined training algorithms or loss functions, and (2) \textbf{Data-centric}:  curated data selection or sophisticated data synthesis. 

\subsection{SFT}
\noindent \textbf{Algorithm-centric SFT.} 
\citet{Li2024EntropicDistributionMatching} introduces \gls{gem}, which reformulates \gls{sft} as a distribution-matching problem with entropy regularization to mitigate overfitting and preserve output diversity. Token Cleaning \citep{Pang2025TokenCleaning} estimates the contribution of each token to model updates and removes uninformative ones, effectively reducing noise in supervision.  One-Token Rollout \citep{Ming2025OneTokenRollout} is a policy-gradient-inspired variant of \gls{sft} that treats each token prediction as a one-step trajectory and leverages the ground-truth token as a reward, introducing on-policy learning signals without the complexity of \gls{rl}.

\noindent \textbf{Data-centric SFT.} 
\citet{zhou2023lima} fine-tunes their model on only $1,000$ high-quality and diverse instruction–response pairs, achieving alignment performance comparable to that of much larger models. FisherSFT \citep{deb2025fishersftdataefficientsupervisedfinetuning} selects training examples that maximize information gain, achieving efficient learning with limited data. \citet{li2025data} optimizes data mixing by learning domain-specific weights that minimize validation loss, thereby improving generalization with minimal tuning cost.  \citet{quan2025automatically} proposes to automatically generate context-driven instruction–response pairs to enrich and diversify \gls{sft} data without heavy human annotation. Finally, Condor \citep{cao2025condor} integrates knowledge-guided synthesis and iterative refinement to produce high-fidelity alignment data, further demonstrating the importance of data curation.

\subsection{RL}
\textbf{Algorithm-centric RL.} Several modified objectives and training procedures have been proposed to extend \gls{llm} capabilities beyond what standard RL achieves. A key line of innovation lies in policy optimization algorithms. While \gls{ppo} \citep{schulman2017proximal}, with a learned value critic, has been the dominant approach, recent work favors critic-free methods such as \gls{grpo} \citep{shao2024deepseekmath}, which simplifies training by replacing the value network with group-relative normalized advantages. Other methods adopt direct REINFORCE-style updates \citep{li2023remax,ahmadian2024back,hu2025reinforce++,xiong2025minimalist}, foregoing the complex components of \gls{ppo}. Another active direction focuses on improving training stability and efficiency through objective regularization. Entropy regularization remains crucial for preventing entropy collapse \citep{cheng2025reasoning,cui2025entropymechanismreinforcementlearning}. \citet{he2025skywork,shrivastava2025sample} introduce weighted token entropy to regularize policy updates. In contrast, \citet{cui2025entropymechanismreinforcementlearning} identifies the covariance between an action’s probability and its advantage as the key entropy “driver,” proposing covariance and \gls{kl} clipping to selectively constrain tokens with exceptionally high covariance. Moreover, \citet{cheng2025reasoning} and \citet{chen2025seed} incorporate entropy information into advantage estimation, further stabilizing training dynamics.

\textbf{Data-centric RL.} Several approaches \citep{zhang2024policy,xu2025not} perform rollout selection, demonstrating that selectively training on a subset of informative model rollouts can achieve promising performance, such as high-variance or diverse examples, hence highly data efficient. Other approaches \citep{zheng2025act,qu2025can} explore prompt selection (prior to rollout) to reduce computational cost while maintaining performance. Curriculum learning also contributes to this direction. \citet{zhang2025clpo,yao2025optimizing} dynamically select prompts of intermediate difficulty to maximize the learning signal. Moreover, \citet{chen2025self, wang2025dump} introduce a distribution-level adaptation approach that prioritizes tasks where the model exhibits the greatest advantage or lowest visitation.

\section{Comparison and Combination of SFT and RL}
\label{sec::theory}

Despite the prevalence of \gls{sft} and \gls{rl}, two commonly adopted post-training techniques for \gls{llm}s, they stay relatively disentangled and are usually applied in a sequential manner. For example, \gls{rlhf} initially applies \gls{sft} to inject prior general knowledge into the \gls{llm}s, and then conducts \gls{rl} for promoting the capability in a particular aspect \citep{ouyang2022training,bai2022training}. However, there is a missing systematic comparison and combination between these two stages, especially from a methodological and theoretical perspective, despite that there have been some pioneering works that try to modify the objectives of each other to complement each other’s strengths. The detailed comparison between different manners to combine \gls{sft} and \gls{rl} together is summarised in Table \ref{table:sft_vs_rl_objective}.



\subsection{A Unified Objective}
First, we write down the objectives of both \gls{sft} and \gls{rl} respectively in an explicit manner,
\begin{align*}
    \cL_{\rm SFT}=\ &\EE_{(x,y)\sim\cD}[-\log\pi_\theta(y|x)],\\
    \cL_{\rm RL}=\ &\EE_{x\sim \cD_x, y\sim \pi_\theta(\cdot|x)}[r(x,y)].
\end{align*}
Based on the above formulation, we could derive the gradients for both objectives below,
\begin{align*}
    \nabla_\theta\cL_{\rm SFT}=\ &\EE_{(x,y)\sim\cD}[-\nabla_\theta\log\pi_\theta(y|x)],\\
    \nabla_\theta\cL_{\rm RL}=\ &\EE_{x\sim \cD_x, y\sim \pi_\theta(\cdot|x)}[\nabla_\theta \log\pi_\theta(y|x)\cdot r(x,y)].
\end{align*}

As pointed out in recent works \citep{wu2025generalization}, the objective of \gls{sft} could be regarded as a special case of \gls{rl}, i.e.,
\begin{align*}
    &\nabla_\theta\cL_{\rm SFT}=\EE_{(x_i,y_i)\sim\cD}[-\nabla_\theta\log\pi_\theta(y_i|x_i)]\\
    &=\EE_{x_i\sim\cD_x}[\EE_{y\sim \pi_\theta(\cdot|x)}[-\frac{\Id\{y=y_i\}}{\pi_\theta(y|x)}\nabla_\theta\log\pi_\theta(y_i|x_i)]],
\end{align*}
where $\Id\{y=y_i\}$ is the indicator function, which takes value $1$ when the output $y$ sampled from policy $\pi_\theta(\cdot|x)$ is exactly the same as the ground-truth response $y_i$ in the \gls{sft} training dataset. Therefore, the ratio $-\frac{\Id\{y=y_i\}}{\pi_\theta(y|x)}$ could be seen as a proxy reward function in \gls{sft}.

Based on the above argument, formulating \gls{sft} as a special case of \gls{rl}, the training objective of most post-training could be abstracted into the following formula,
\begin{align*}
    \max_{\pi}&\EE_{x\sim\cD_x,y\sim\pi_\theta(\cdot|x)}\big[r(x,y)\big] \\ &-\beta\ {\rm KL}\big(\pi_\theta(\cdot|x)\|\pi_0(\cdot|x)\big),
\end{align*}
where $\pi_0$ is the base (reference) policy model; $r$ is a proxy for the reward, and $\beta$ is a hyper-parameter. The \gls{kl} regularization between the policy model and the reference model restricts $\pi_\theta$ from deviating too much from a pre-trained checkpoint, primarily for stability reasons. In the following sections, we will continue to discuss the difference and combination of \gls{sft} and \gls{rl}, mainly from the objective perspective.

Therefore, optimizing \gls{llm}s with both \gls{sft} and \gls{rl} objectives ultimately collapses to the \gls{rl} objective, and the tricks that work for one thus have potential to be applied to the other. To mitigate the drop in generalization ability from \gls{sft}, one could consider applying importance sampling and online rollout, as in \gls{rl}. Meanwhile, to help \gls{llm}s memorize additional knowledge, one could integrate the \gls{sft} loss into the \gls{rl} objective. The roles of these two stages should be regarded as a mutually reinforcing and interdependent relationship instead of merely being applied alternatively.

\subsection{Leveraging SFT to Enhance RL}
\paragraph{Combine offline demonstration and online data} Recent studies have leveraged \gls{sft} to enhance \gls{rl} in \gls{llm}s by combining offline demonstrations with online rollouts. \citet{yan2025learning} introduces an off-policy guided framework that augments on-policy updates with reasoning traces, effectively balancing imitation and exploration. SRFT \citep{fu2025srft} integrates supervised and reinforcement objectives in a single-stage framework, avoiding inefficiencies of sequential fine-tuning. Similar to \gls{llm}s, \gls{vlm} are another flourishing area with great potential by extending the single text-based modality of \gls{llm}s into vision, like images and videos. \citet{liu2025empowering} propose a dynamic memorization--exploration strategy for small \gls{vlm}, which adaptively alternates between demonstration imitation and online reward optimization. AMFT \citep{he2025amft} adopts a meta-learning perspective, dynamically adjusting the imitation--exploration balance to maximize performance. \citet{lv2025towards} provides a unified view of post-training, showing that both \gls{sft} and \gls{rl} optimize a common objective, and propose hybrid updates to exploit demonstration and rollout data. BREAD \citep{zhang2025bread} bridges \gls{sft} and \gls{rl} through branched rollouts anchored by expert prefixes, reducing reliance on large demonstration sets while improving stability. Similarly, \citet{huang2025blending} develops a prefix sampling approach that guides generation into high-quality trajectories before applying \gls{rl} optimization. Collectively, these methods illustrate complementary strategies, including single-stage integration, adaptive weighting, and prefix-based seeding, that enhance the efficiency and robustness of \gls{rl} when grounded in \gls{sft} demonstrations. 

\paragraph{Objective Modification}
NFT \citep{chen2025bridging} enables models to learn from both correct and incorrect outputs under supervision while implicitly optimizing a policy to enhance reasoning performance. UFT \citep{liu2025uft} unifies supervised and reinforcement fine-tuning into a single process that balances memorization and exploration, achieving better generalization and exponentially improved sample efficiency. ReLIFT \citep{zhu2025surprising} interleaves \gls{rl} with \gls{sft} on the hardest questions, allowing models to acquire capabilities beyond what pure \gls{rl} can achieve. \citet{zhu2025surprising} demonstrates that penalizing incorrect answers alone, through negative sample reinforcement, can substantially improve reasoning performance, often matching or exceeding classical \gls{rl} methods by suppressing wrong outputs and reallocating probability to plausible alternatives.

\subsection{From RL Perspective to Improve SFT}
DFT \citep{wu2025generalization} is proposed to rescale each token’s loss by its predicted probability to rectify implicit reward bias in \gls{sft} and boost generalization.  
iw-SFT \citep{qin2025supervised} interpret curated \gls{sft} as optimizing a lower bound on a sparse-reward \gls{rl} objective and tightening that bound via importance weights.  
Another work \citep{du2025simplify} in \gls{rlhf} proposes a reweighted reward-driven \gls{sft} objective through variational inference. In contrast, inspired by the \gls{rl}, especially \gls{dpo} \citep{rafailov2023direct}, objective, \citet{wang2024uft} minimizes the distribution difference between the reference model and policy model on an offline dataset.

\subsection{Hybrid Training Combining SFT and RL}
\citet{huang2025blending} proposes to combine \gls{sft} and \gls{rl} together through utilizing the prefix of a ground-truth response to generate new continuation based on the current policy model, and use \gls{sft} loss to train the prefix partition while use \gls{rl} loss to train the newly generated partition. \citet{lv2025towards} uses interleaving \gls{sft} and \gls{rl} based on the performance of the policy model during online rollouts. When the performance is above a preset threshold, online \gls{rl} is preferred for exploration, while correct guidance from \gls{sft} is preferred when the performance is bad. \citet{liu2024provably} starts from the original \gls{dpo} \citep{rafailov2023direct} objective, and directly injects the \gls{sft} loss on samples from the base model to mitigate the overoptimization. \citet{zhang2025policy} combines a weighted \gls{sft} loss and \gls{grpo} \citep{shao2024deepseekmath} loss together to aggregate the off-policy and on-policy training. \citet{liu2025uft} uses expert data in the first several steps, and switches to the new generation in the following steps. SRL \citep{deng2025supervised} proposes to decompose the reasoning process into several intermediate steps, and compare online rollouts with offline expert trajectories, and assign rewards proportional to the matching between them.

\begin{sidewaystable*}[h]
\centering
\caption{Objective comparison across training paradigms. Here for integration, (SFT $\rightarrow$) RL means modifying RL objective based on SFT objective, and vice versa. 
}
\label{table:sft_vs_rl_objective}
\scriptsize
\renewcommand{\arraystretch}{3.0}
\begin{tabularx}{\textwidth}{@{}m{0.1\textwidth}m{0.14\textwidth}m{0.47\textwidth} m{0.12\textwidth} m{0.14\textwidth}@{}}
\toprule
\midrule
\textbf{Method} & \textbf{Paper} & \textbf{Objective / Gradient Expression} & \textbf{Data} & \textbf{Integration} \\
\midrule
DFT & \citet{wu2025generalization} &
$\EE_{(x_i,y_i)\sim\cD}[-{\rm sg}(\pi_\theta(y_i|x_i))\nabla_\theta \log\pi_\theta(y_i|x_i)]$ & Offline dataset & (RL$\rightarrow$) SFT \\
Prefix Sampling & \citet{huang2025blending} & \makecell[l]{
$-\sum_{y_1,\cdots,y_{n-1},y_n^{<L}}\alpha\nabla_\theta\log\pi_\theta(y_i^{t}|x,y_i^{<t})$\\
\qquad $-\sum_{y_n^{<L}}\beta\nabla_\theta\log\pi_\theta(y_n^t|x,y_n^{<t})$
} & Offline dataset + online rollout & RL + SFT \\
CHORD & \citet{zhang2025policy} & \makecell[l]{$\cL_{\rm CHORD}(\theta)=(1-\mu)\cL_{\rm GRPO}(\theta)+\mu\cL_{\rm SFT-\phi}(\theta)$, where\\ $\cL_{\rm SFT-\phi}=-\mathbb{E}_{(x, y^*) \sim \mathcal{D}_{\text{SFT}}}
[\sum_{t=1}^{|y^*|}\phi(y_t^*; \pi_\theta)
\cdot \log \pi_\theta(y_t^* \mid x, y_{<t}^*)]$, and\\ $\phi(y_t^*;\pi_\theta)=\pi_\theta(y^*_t|x,y^*_{<t})(1-\pi_\theta(y^*_t|x,y^*_{<t}))$} & Offline dataset + online rollout & RL + SFT \\
UFT & \citet{liu2025uft} & \makecell[l]{$\mathcal{J}^{\mathrm{UFT}} = \mathbb{E}_{\substack{l, x_l = x_l^* \\ (x_h, y_h)_{h = l}^{H-1} \sim \pi}}\![\mathcal{J}^{\mathrm{value}}\!\big((x_h, y_h)_{h = l}^{H-1}\big)$ \\\qquad
$ - \beta \!\sum_{h = l}^{H-1} \mathrm{KL}\!\big(\pi(\cdot \mid x_h)\,\|\,\pi^{\mathrm{ref}}(\cdot \mid x_h)\big) + \beta \!\sum_{h = 0}^{l-1} \log \pi(y_h^* \mid x_h^*)]$}
& Offline dataset + online rollout & RL + SFT \\
Proximal SFT & \citet{zhu2025proximal} & $L^{\mathrm{PSFT}}(\theta)
= \mathbb{E}_{(x_t, y_t) \sim \mathcal{D}}
\!\left[
\min\!\left(
\frac{\pi_\theta(y_t \mid x_t)}{\pi_{\theta_{\text{old}}}(y_t \mid x_t)},
\operatorname{clip}\!\left(
\frac{\pi_\theta(y_t \mid x_t)}{\pi_{\theta_{\text{old}}}(y_t \mid x_t)},
1 - \epsilon,\,
1 + \epsilon
\right)
\right)
\right] $ & Offline dataset & (RL $\rightarrow$) SFT \\
UFT-SFT & \citet{wang2024uft} & $L_{\mathrm{UFT\text{-}SFT}}(\pi_\theta)
= \mathbb{E}_{(x,y)\sim\mathcal{D}}
\!\left\{
\left[
\sigma\!\left(
\beta \log\!\frac{\pi_\theta(y|x)}{\pi_{\mathrm{ref}}(y|x)}
\right) - 1
\right]^2
\right\}$& Offline dataset & (RL $\rightarrow$) SFT \\
iw-SFT & \citet{qin2025supervised} & $\cL_{\rm iw-SFT}=\EE_{\tau\in\cD^+}[-\frac{q(\tau)}{\pi_{\rm ref}(\tau)}\log p(\tau;\theta)]$ & Offline dataset & (RL $\rightarrow$) SFT \\
VAR & \citet{du2025simplify} & $\cL=-\EE[\frac{\pi_{\rm ref}(y|x)\exp(\frac{1}{\lambda}r(x,y)}{Z(x)}\log\pi_\theta(y|x)]$ & Offline dataset & SFT \\
HPT & \citet{lv2025towards} & $\cL=\alpha\cL_{\rm RL}+\beta\cL_{\rm SFT}$ & Offline dataset + online rollout & RL + SFT\\
RPO & \citet{liu2024provably} & \makecell[l]{$\cL_{\rm RPO}(\theta)=\eta\beta\cdot \EE_{x\sim d_0,y^0\sim\pi^{\rm base}(\cdot|x)}[-\log\pi_\theta(y^0|x)]+\cL_{\cD}(\beta\cdot\log\frac{\pi_\theta}{\pi_{\rm ref}})$,\\ where $\mathcal{L}_{\mathcal{D}}(r)
= -\,\widehat{\mathbb{E}}_{\mathcal{D}}\!\left[
p_i \log\!\big(\sigma(r(x_i,y_i^{1}) - r(x_i,y_i^{0}))\big)
+ (1 - p_i)\log\!\big(\sigma(r(x_i,y_i^{0}) - r(x_i,y_i^{1}))\big)
\right]$} & Offline (preference) dataset & SFT + DPO \\
SRFT & \citet{fu2025srft} & \makecell[l]{$\cL_{\rm SRFT}(\theta)=\cL_{\rm SFT}^{\rm demo}(\theta)+\cL_{\rm RL}^{\rm demo}(\theta)+\cL_{\rm RL}^{\rm self-rollout}(\theta)$, where $\cL_{\rm RL}^{\rm self-rollout}$ stands for SFT loss on\\ new rollouts from policy model, and GRPO advantage estimation is used in $\cL_{\rm RL}^{\rm demo}$ with concatenated group from\\ offline dataset and online rollout} & Offline dataset + online rollout & SFT + RL \\
Q-SFT & \citet{hong2024q} & \makecell[l]{$\cL_{\rm CE}(\phi)=\EE_{(x,y)\sim\cD}[\log\pi_\phi(y|x)]$ \\ $\mathcal{L}_{\mathrm{TD}}(\theta)
= \mathbb{E}_{(x,y,r,x^\prime)\sim\mathcal{D}}
\!\left[
\big(
r + \gamma \max_{y^\prime} Q_{\bar{\theta}}(x^\prime,y^\prime) - Q_{\theta}(x,y)
\big)^2
\right]$} & Offline dataset & SFT + Q learning \\
IRL-RFT & \citet{li2024getting} & \makecell[l]{$\frac{1}{\beta}\nabla_\theta r(x_{t,k},y_{t,k};\theta_{t,k})-\frac{1}{\beta}\nabla_\theta r(x_{t,k},\tilde{y}_{t,k};\theta_{t,k})$, where $y\sim\pi_{\rm expert}(\cdot|x_{t,k})$,\\ $\tilde{y}\sim\pi^t(\cdot|x_{t,k}),\ \pi^t(y|x)\propto \exp(r(x,y;\theta_{t,K}))$} & Online rollout & RL \\
SASR & \citet{chen2025step} & $\cL(\theta)=\frac{1}{S}\sum_{s=1}^S[(1-I(t))\cdot \cL_{\rm SFT}(\theta)+I(t)\cdot \cL_{\rm GRPO}(\theta)]$ & Offline dataset + online rollout & SFT + RL \\
IFT & \citet{hua2024intuitive} & $\mathcal{L}\!\left(\hat{T}_{\theta}(s_n^{*}, \rho_0)\right)
= -\sum_{i=n}^{N} \log \mathcal{T}_{\theta}\!\big(a_i^{*}, \delta_{\theta}(s_i^{*})\big)$ & Offline dataset & RL \\
NFT & \citet{chen2025bridging} & \makecell[l]{$\mathcal{L}_{\mathcal{D}}^{\mathrm{NFT}}(\theta)
= -\!\!\sum_{x,y,r}\!\omega(x)\!\sum_{t}
\!\left[
r\log R_{\theta}^{t}(x,y)
+ (1 - r)\log\max_{v}\!\left(\frac{1-\hat{r}_{x}\,R_{\theta}^{t}(x,y)}{1-\hat{r}_{x}},\,\epsilon\right)
\right]$, where\\ $R_{\theta}^{t}(x,y)
= \frac{\pi_{\theta}^{+}(y_{t}\mid x,y_{<t})}{\pi(y_{t}\mid x,y_{<t})},
\
\hat{r}_{x}
= \frac{1}{K}\!\sum_{y\mid x}r(x,y)$} & Online rollout & (SFT $\rightarrow$) RL \\
SRL & \citet{deng2025supervised} & \makecell[l]{ $R=\frac{2\sum_{(i,j,n)\in{\rm MatchingBlocks}}n}{|S_1|+|S_2|}$, $r(\mathbf{y}'_{\text{step}_{p_k}}, \mathbf{y}_{\text{step}_{p_k}}) = 
\begin{cases}
R(\mathbf{y}'_{\text{step}_{p_k}}, \mathbf{y}_{\text{step}_{p_k}}), & \text{if } \mathbf{y}' \text{ follows format},\\
-1, & \text{otherwise.}
\end{cases}$} & Offline dataset + online rollout & SFT + RL \\
\bottomrule
\end{tabularx}
\end{sidewaystable*}


\section{Applications}
\label{sec:application}

\begin{figure*}[ht] 
    \centering
    \includegraphics[width=0.85\textwidth]{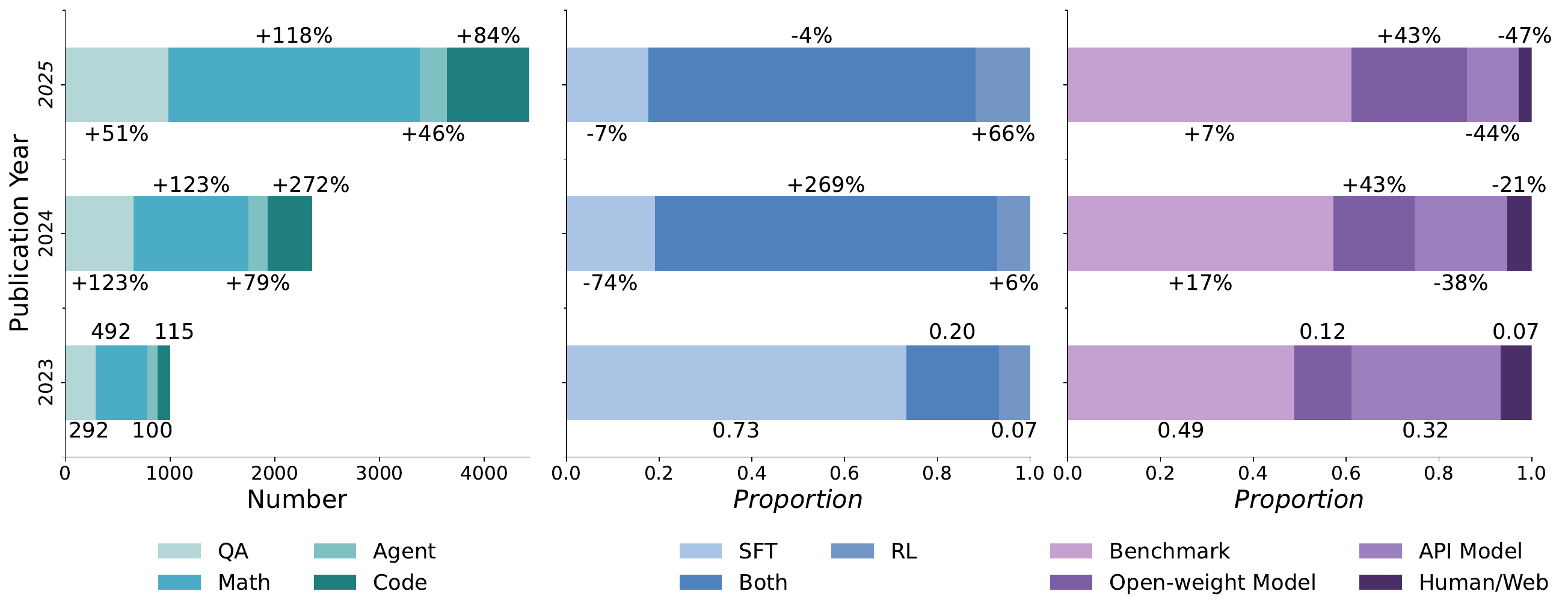}
    \caption{Trends in task focus, training methodologies, and ground-truth data sources from 2023 to 2025. Results show rapid growth across all surveyed domains, with substantial increases in research volume and diversification of application areas; increasing convergence toward hybrid SFT–RL pipelines, supported by more mature training infrastructures, libraries, and preference datasets; and a continued shift from API-based labeling to data generated with increasingly capable open-weight models. Projections for 2025 and all reported proportions are derived from surveyed publications; see Appendix \ref{appendix:fig_dis} for further discussion.}
    \label{fig:app}
\end{figure*}

We focus on application studies that post-train \gls{llm}s using \gls{sft} or \gls{rl} with four domains, where each domain encompasses distinct methodologies and challenges, and these categories capture the breadth of \gls{llm} capabilities and support a systematic analysis of performance across tasks. Key trends are summarized in Fig. \ref{fig:app}, with additional details provided in Appendix \ref{appedix:application}.

\subsection{LLMs for General QA Tasks}
This subsection reviews methods for improving \gls{llm} \gls{qa} performance via reasoning augmentation, e.g., \gls{cot}; external knowledge integration, e.g., \gls{rag}; and hallucination mitigation techniques.

\textbf{Step-by-Step Reasoning.}
Smaller \gls{fm} trained only on web data generally lack native \gls{cot} reasoning; thus, \gls{cot} demonstrations are typically generated using larger models via prompt engineering and subsequently used for \gls{sft} to instruction-tune reasoning or query decomposition capabilities.
Building upon this approach, some works \citep{qa_9, qa_11} introduce an additional \gls{rl} stage, where collected demonstrations are treated as positive examples and newly generated model outputs are treated as negative samples.
A key challenge in equipping \gls{llm}s with \gls{cot} capabilities lies in evaluating the quality of generated reasoning, as providing fine-grained, step-wise supervision in an online setting is often difficult or costly. To mitigate this, researchers have proposed using surrogate signals, such as answer correctness \citep{qa_4, deepseek2024r1} or reasoning comprehensiveness \citep{qa_14}, and applying \gls{rft} or \gls{rl} on self-generated samples.

\textbf{LLM-based Retrieval.}
In \gls{rag} pipelines, \gls{llm}s are often used for query rewriting prior to retrieval. Prior work shows that \gls{fm}s can perform this task effectively, with retrieval quality evaluated via downstream responses, and several studies \citep{qa_10, qa_16, qa_18, qa_21, qa_15} directly apply \gls{rl} with task-specific search rewards for model alignment.

\textbf{Hallucination Management.}
\gls{llm} hallucination refers to the generation of responses that appear plausible but contain incorrect or fabricated information. This phenomenon typically arises when the required factual knowledge is absent from the pre-training corpus, leading the model to infer or construct an answer rather than abstain.
To mitigate hallucinations for general question, researchers have leveraged \gls{ppo} to reduce factual errors by incorporating response corrections \citep{qa_2} provided by stronger models. Additionally, to encourage models to explicitly acknowledge uncertainty and suppress hallucinated outputs, prior works \citep{qa_6, qa_1, qa_3, qa_7} have proposed inference over a collection of sampled responses, often derived from benchmark datasets, and synthesizing “I do not know” outputs for subsequent \gls{sft} or \gls{rl} pipelines.
Furthermore, to quantify and model uncertainty, several studies 
\citep{qa_8, qa_20} generate multiple candidate responses for each training instance and fine-tune the model using aggregated or summarized outputs as approximations of the underlying uncertainty distribution.

\subsection{LLMs for Mathematical Tasks}
This subsection reviews methods for enhancing \gls{llm}s' mathematical capabilities, with a focus on joint mathematical reasoning and training using step-wise rollouts.

\textbf{Mathematical Reasoning.}
Early efforts for mathematical reasoning abilities can be traced back to \citet{math_data_1} and \citet{math_verify}, which studied problems involving algorithmic computation and mathematical induction.
Subsequent work further improved performance by \gls{sft} \gls{fm}s on curated mathematical corpora collected from the web \citep{math_data_3, math_data_2}, synthetically generated datasets \citep{math_evol_data_2, math_evol__1}, or data produced by stronger models \citep{math_evol_data_1, math_evol__0}.
More recent studies construct explicit reward models for mathematical evaluation and propose alignment procedures, such as \gls{grpo} \citep{shao2024deepseekmath} and Critique-DPO \citep{math_verify_6}, to enable downstream \gls{rl}.

\textbf{Rollout-Based Training.}
Similar to standard \gls{cot} prompting, developing reliable step-wise reward evaluation for mathematical reasoning remains challenging.
\citet{math_verify_4} proposes fine-tuning models on their own generated solutions using majority voting.
Additionally, many researchers exploited \gls{fm}s with multiple reasoning paths when solving mathematical problems.
For \gls{ppo}-based methods, \citet{math_verify_2} and \citet{math_verify_3} approximate process-level rewards using the probability of producing a correct final answer, while \citet{math_RL_step} randomly assigns reflection prompts and incorporates both outcome-based and reflection-based bonus rewards.

\subsection{LLMs for Agentic Tasks}
This subsection reviews strategies for improving the agentic capabilities of \gls{llm}s, with a focus on action selection and long-horizon planning.

\textbf{Action Selection.} \label{app:agent1}
Agentic tasks require selecting actions over multiple steps within a given environment.
To equip \gls{fm}s with this capability, pioneering work such as \citet{schick2023toolformer, sft_data_3, sft_data_4, sft_data_0} synthesizes or unifies action trajectories across diverse environments for \gls{sft}. 
Subsequent studies adopt a sequential paradigm that combines \gls{sft} with \gls{rl} using environment-specific rewards.
For example, \citet{both_sametime_2} incorporates \gls{sft} rewards directly into the \gls{sft} objective, while hierarchical \gls{rl} approaches \citep{rl_alg_1, rl_alg_6} assign rewards at the sentence level while updating policies at the token level.
Similarly, DMPO \citep{rl_alg_3} introduces step-wise reward discounting, and \citet{rl_alg_0} employs critic-based curriculum learning to stabilize \gls{rl} training.

\textbf{Planning with LLMs.}
Task planning has emerged as an active research direction, with agentic tasks serving as a natural testbed due to their long action horizons and complex inter-step dependencies.
Recent work predominantly leverages outcome-based surrogate rewards and GPT-synthesized sequences to optimize planning behavior.
For instance, \citet{both_seq_3} proposes a hierarchical planning framework with a top-level planner fine-tuned on seed plans and optimized via \gls{dpo}.
\citet{both_seq_4} further incorporates \gls{rft} to refine high-reward actions, while \citet{rl_data_4} improves planning performance by adjusting exploration strategies and prioritizing critical steps during \gls{dpo} training.

\subsection{LLMs for Coding Tasks}
This subsection reviews methods for improving coding capabilities in \gls{llm}s, focusing on code generation and code editing.

\textbf{Code Generation.}
Recent work has explored various approaches to enhance \gls{llm}-based code generation.
Under \gls{sft}, \citet{code_gen_1} employs self-instruction to synthesize training data, while \citet{code_gen_2} generates more challenging queries by modifying constraints of seed problems.
\citet{code_gen_3} leverages open-source code snippets to produce diverse and controllable instruction datasets.
From the \gls{rl} perspective, \citet{code_gen_4} introduces a multi-agent framework in which an \gls{sft} model generates rollouts for subsequent \gls{dpo} training.

\textbf{Code Editing.}
Code editing aims to refine programs that fail to satisfy intended functionality, whether human-written or \gls{llm}-generated.
For example, \citet{code_edit_6} leverages human pull requests for \gls{grpo}-based training, while \citet{code_edit_4} and \citet{code_edit_2} generate iterative refinement rollouts evaluated using unit tests to drive \gls{rl} optimization.

\section{Common Practice and Takeaways}
In this section, we summarize common practices and key takeaways from our previous theoretical and application discussions, highlighting the interplay between \gls{sft} and \gls{rl}, when to choose each method, and training trade-offs.

\textbf{Relationship Between SFT and RL Objectives.}
As discussed in Section \ref{sec::theory}, \gls{sft} can be viewed as a special case of \gls{rl} under certain formulations, where an indicator function that measures whether the current policy reproduces an offline trajectory acts as a surrogate reward.
This shared optimization structure enables techniques developed for either \gls{sft} or \gls{rl} to transfer across paradigms.
For example, Section \ref{app:agent1} reviews several studies that jointly optimize weighted \gls{sft} and \gls{rl} objectives, demonstrating consistent improvements over standalone training losses.

\textbf{Leveraging Expert Data or Policies.}
When high-quality expert data are available, \gls{sft} is generally preferred as an initial training stage over RL as observed in Section \ref{sec:application}, possibly due to its simpler implementation and greater stability. 
If a strong policy model or prompt is available to generate expert data, incorporating importance sampling on failed queries can mitigate distribution shift by treating them as informative positive samples for \gls{rft}/\gls{rl}, bridging the gap between the limited offline dataset and the evolving policy. 
When a reliable reward model can be trained, the common practice is to first perform \gls{sft} and then \gls{rl}, which typically achieves the highest reported performance.

\textbf{Training Strategies and Trade-offs.}  
\gls{sft} is essential when the policy model is unfamiliar with the downstream task or cannot efficiently generate sufficient positive samples. However, \gls{sft} may reduce the model's generalization ability compared with \gls{rl}. Conversely, directly applying \gls{rl} can better balance performance gains with exploratory capacity, though it may suffer from issues such as entropy collapse \citep{cui2025entropymechanismreinforcementlearning} or reward hacking \citep{pan2024feedbackloopslanguagemodels, pan2024spontaneousrewardhackingiterative}, one possible reason why the majority of pre-training computation is still devoted to \gls{sft}.

\section{Future Directions and Open Problems}
\label{app:future}
Despite rapid progress, many fundamental questions remain unresolved. We highlight two open challenges and outline promising directions for future research.

\textbf{Sample- and compute-efficient methodologies.}
State-of-the-art \gls{sft} and \gls{rl} pipelines typically require substantial computational resources, large volumes of high-quality data, and extensive rollout generation, raising both practical and environmental concerns. Improving efficiency is therefore a central challenge. Early progress includes data-efficient \gls{sft} based on information-theoretic principles \citep{deb2025fishersftdataefficientsupervisedfinetuning} and quantization-aware methods that reduce training cost \citep{wei2025rosteefficientquantizationawaresupervised}. For \gls{rl}, recent work explores selective or partial rollouts to reduce computation \citep{zheng2025act, xu2025rolloutsusefuldownsamplingrollouts}. Nevertheless, significantly more research is needed to develop broadly applicable, highly efficient methods.

\textbf{SFT and RL under sparse or indirect reward signals.}
Many real-world tasks lack well-defined or easily verifiable reward signals. User feedback may be sparse, inconsistent, or costly. In safety-critical settings, it may be infeasible to obtain ground-truth evaluations altogether. Recent work has begun to explore alignment through verbal or otherwise indirect feedback \citep{stephan2024rlvflearningverbalfeedback, both_sametime_1}. An open challenge is to identify and leverage additional natural but underexplored sources of implicit supervision, such as self-evaluation signals or user-churn behaviors, to guide both \gls{sft} and \gls{rl} in low-feedback regimes.

\section{Conclusion}
This study provides a unified view of \gls{sft} and \gls{rl} as post-training strategies for \gls{llm}s, examining both their theoretical foundations and practical implementations. \gls{sft} offers stable and efficient learning from high-quality offline datasets, whereas \gls{rl} enables reward-driven optimization that enhances generalization and exploration. Recent research characterizes \gls{sft} as a special case of \gls{rl}, underscoring the promise of hybrid approaches that balance training stability with adaptive behavior. Application papers published between 2023 and 2025 indicate a clear shift toward hybrid post-training pipelines that integrate \gls{sft} and \gls{rl}, with scalable workflows centered on open-weight, model-generated rollouts emerging as the dominant paradigm. These developments underscore the need for integrative methodologies that leverage the strengths of both approaches, and this study lays a foundation by consolidating the current state of the field, theoretical advances, and versatile \gls{llm} applications.

\section*{Limitations}
Despite our best efforts to ensure comprehensiveness, this study may omit some recent advances in SFT/RL research due to the rapid pace of progress and the sheer volume of emerging literature in this field. In the \textit{Applications} section, the use of a paper-filtering strategy may introduce approximation bias, as commonly adopted benchmarks are not fully inclusive of all real-world scenarios or methodologies. Additional discussion of application approximation limitations is provided in Appendix \ref{app:b2}.


\bibliography{main}

\newpage
\appendix
\clearpage

\section{More on Comparing and Combining SFT and RL}
We summarize the main notation used throughout this section and Table~\ref{table:sft_vs_rl_objective}. 
For ease of reference, commonly used acronyms are listed in the acronym table, and a consolidated summary of symbols is provided in Table~\ref{tab:symbol_summary}.

\subsection{Notation}
The offline training dataset used in SFT and related settings is denoted by $\mathcal{D}=\{(x_i, y_i)\}$, where $x_i$ represents the input prompt and $y_i$ denotes the corresponding model-generated response. 
We use $\pi_\theta$ to denote a policy (language) model parameterized by $\theta$, and $\pi_\theta(y_i \mid x_i)$ to denote the conditional probability of generating response $y_i$ given prompt $x_i$. 
The reference policy is denoted by $\pi^{\rm ref}$, which is typically a frozen base model or an SFT checkpoint.

In the RL setting, $r(x_i, y_i)$ denotes the reward assigned to a prompt-response pair $(x_i, y_i)$, and $Q_\theta(x, y)$ represents the action-value (Q) function under policy $\pi_\theta$, with state $x$ (prompt) and action $y$ (response). 
We use $\mathcal{J}$ to denote cumulative reward–based objectives.

The sigmoid function is defined as $\sigma(x) = 1/(1+e^{-x})$. 
The operator ${\rm sg}(\cdot)$ denotes the stop-gradient operation, which preserves the forward-pass value while preventing gradient propagation during backpropagation.
Additional symbols, operators, and hyperparameters appearing in Table \ref{table:sft_vs_rl_objective} are summarized in Table \ref{tab:symbol_summary} for completeness.

\subsection{Preliminaries}
We consider the optimization of Pretrained LLMs, also referred to as FMs, which are large-scale pretrained models adapted to downstream tasks such as QA, reasoning with CoT, and database augmentation tasks such as RAG. 
Training typically proceeds in multiple stages, starting with SFT on curated prompt-response pairs, which can be viewed as a form of BC from expert demonstrations.

Beyond SFT, model alignment and performance are often further improved using RL techniques, particularly RLHF, where a learned or human-provided reward signal guides optimization.
Popular policy-gradient–based algorithms include PPO and its variants such as GRPO, which stabilize updates through clipping or relative normalization.
Alternative optimization frameworks include preference-based methods such as DPO and rejection-based methods such as RFT, which bypass explicit reward modeling.
Information-theoretic objectives, such as GEM, and regularization via the KL divergence are commonly used to control deviation from a reference policy.

The RL process itself can be formalized as a MDP, defined by the tuple $(\mathcal{S}, \mathcal{A}, P, r, H)$, where $\mathcal{S}$ is the state space, $\mathcal{A}$ is the action space, $P$ denotes the transition dynamics, $r$ is the reward function, and $H$ is the horizon length, which may be infinite under idealized analysis.
The policy $\pi_\theta(\cdot \mid x)$ represents an LLM that induces a distribution over responses conditioned on an input prompt $x$.
This formulation naturally extends to multi-modal settings, including VLM, and other structured decision-making scenarios.

\printglossary[type=\acronymtype,title={Acronyms from the Main Paper}]

\begin{table}[H]
\centering
\begin{tabularx}{\columnwidth}{@{}l X@{}}
\toprule
\textbf{Symbol} & \textbf{Description} \\
\midrule
\multicolumn{2}{l}{\textit{Models and Policies}} \\
$\pi_\theta, \pi_\phi$ & Current policy/model parameterized by $\theta$ or $\phi$ \\
$\pi_{\text{ref}}, \pi_{\text{base}}$ & Reference or base policy (typically frozen) \\
$\pi_{\text{old}}$ & Policy from a previous optimization step \\
$\pi_{\text{expert}}$ & Expert policy used for imitation or IRL \\
\midrule
\multicolumn{2}{l}{\textit{Data and Sequences}} \\
$x, y$ & Input prompt and generated response \\
$y_t, y_{<t}$ & Token at time $t$ and the prefix sequence \\
$\mathcal{D}, \mathcal{D}^+$ & Dataset (offline, SFT, or preference-based) \\
$\tau$ & Trajectory or full sequence $(x, y)$ \\
\midrule
\multicolumn{2}{l}{\textit{Functions and Objectives}} \\
$\mathcal{L}, \mathcal{J}$ & Loss or objective function \\
$r(x, y), R$ & Reward function or scalar reward value \\
$Q(x, y)$ & Action-value function (Q-function) \\
$Z(x)$ & Partition function (normalization constant) \\
$\sigma(\cdot)$ & Sigmoid activation function \\
\midrule
\multicolumn{2}{l}{\textit{Operators and Hyperparameters}} \\
$\mathbb{E}$ & Mathematical expectation \\
$\nabla_\theta$ & Gradient with respect to parameters $\theta$ \\
$\text{sg}(\cdot)$ & Stop-gradient operator \\
$\text{KL}(\cdot \| \cdot)$ & Kullback-Leibler divergence \\
$\alpha, \beta, \mu, \eta$ & Weighting coefficients or temperature parameters \\
$\epsilon$ & Clipping or threshold constant \\
\bottomrule
\end{tabularx}
\caption{Summary of Notation and Symbols}
\label{tab:symbol_summary}
\end{table}

\section{Application Study Details}
\label{appedix:application}
This appendix section provides detailed information on our study of LLM applications across professional domains, including the datasets, evaluation criteria, and publication trends that underpin our analysis. The goal is to contextualize the results presented in the main text by describing (i) how different task domains vary in input complexity, reasoning demands, and output characteristics, (ii) our methodology for categorizing research publications by domain using benchmark-oriented searches, and (iii) observed trends in publication volume, methodological approaches, and model usage over time. By presenting these details, we aim to offer a transparent and reproducible account of the study process, enabling deeper insight into the evolving landscape of LLM applications.

\subsection{Applications Across Professional Domains}
To characterize the difficulty of each domain, we consider three qualitative factors: input complexity, intermediate reasoning requirements (chains of thought, CoT), and output complexity. Table \ref{tab:domain_summary} summarizes these characteristics across focompur representative domains. Note that Section \ref{sec:application} focuses on papers addressing text-only tasks. We make this choice because text-only tasks enable clearer side-by-side comparison, whereas multimodal settings introduce additional sources of uncertainty in model selection and fusion techniques, which lie outside the scope of this work. Nevertheless, we acknowledge that numerous impactful works closely related to the studied area extend beyond text-only modalities. For further references, we direct readers to Section \ref{appendix:more_ref}, which highlights some recently published studies as a starting point.

General QA assesses fact-checking and language understanding with variable-length inputs and moderate reasoning, prioritizing accuracy over verbosity. Mathematical tasks demand precise answers supported by extended logical reasoning. Agentic tasks require rich environmental context to enable multi-step planning before producing compact outputs. Code generation relies on context-heavy inputs and moderate reasoning to infer intent and produce syntactically correct, semantically coherent code.


\begin{table*}
\centering
\caption{Summary of LLM application domains and their characteristics.}
\label{tab:domain_summary}
\begin{tabular}{lccc}
\toprule
\textbf{Domain} & \textbf{Input} & \textbf{CoT / Reasoning} & \textbf{Output} \\
\midrule
General QA      & Moderate & Moderate                 & Moderate \\
Mathematical    & Moderate & Long                     & Moderate \\
Agentic         & Long     & Long(Multi-Step)         & Moderate \\
Code Gen.       & Long     & Moderate                 & Long \\
\bottomrule
\end{tabular}
\end{table*}

Across these domains, the overarching question is how to optimize performance given varying input scales, reasoning depths, and output complexities. The relative weight of these factors differs by task, reflecting the diverse requirements and shifting priorities of professional LLM applications.

\subsection{Benchmark-Oriented Paper Search} \label{app:b2}
To estimate publication counts for each domain, we conducted a benchmark-oriented keyword search over arXiv preprints in the Computer Science categories from 1 January 2023 to 30 June 2025. The arXiv metadata dataset (Cornell-University/arxiv) is provided under the Creative Commons CC0 1.0 Universal Public Domain Dedication, which applies to the dataset’s metadata as hosted on Kaggle. Original arXiv content remains subject to the licenses specified by individual authors. We accessed scientific articles via Google Cloud Public Datasets (\url{gs://arxiv-dataset}). The initial dataset comprises approximately 195K papers, with 64K, 78K, and 52K papers from 2023, 2024, and 2025, respectively. Our approach is motivated by the observation that recent LLM post-training research consistently relies on a set of widely used, standardized datasets that facilitate reproducible and comparable evaluations across methods. As this practice has become increasingly common, dataset mentions serve as a reasonable proxy for assigning papers to topical domains.

Following Table 7 of \citet{du2025survey}, we use 26 datasets spanning four domains as domain-specific query keys. A paper is assigned to a domain if it contains at least five mentions of any dataset associated with that domain. Mutual exclusivity is not enforced: although relatively rare, papers that evaluate across multiple domains are counted in every applicable category.

The datasets used as search keys are listed in Table \ref{tbl::app::key}, and the resulting paper counts for each category are shown in Table \ref{tab:threshold}. We adopt a threshold of five keyword occurrences because a paper that genuinely employs a dataset typically introduces it, reports results on it, and provides comparative or analytical discussion, making five mentions a conservative filter against purely methodological contributions. To project the full-year counts for 2025, we note that the first six months of 2023 and 2024 account for 47.38\% and 49.81\% of their respective annual publication totals. Accordingly, we estimate the 2025 full-year count by doubling the observed number from the first half of 2025.

\begin{table*}
\centering
\caption{Datasets used as search keys for benchmark-oriented paper classification.}
\begin{tabular}{l}
\toprule
\textbf{Agentic} \\
webshop, webarena, wind2web, miniwob++, scienceworld, alfworld, \\
tdw-mat, c-wah, alfred, rlcard \\
\midrule
\textbf{QA} \\
hotpotqa, strategyqa, triviaqa, pubmedqa, musique, \\
2wikimultihopqa, qasper \\
\midrule
\textbf{Math} \\
gsm8k, asdiv, svamp, aime \\
\midrule
\textbf{Code} \\
swe-bench, humaneval, livecodebench, bird, intercodesql \\
\bottomrule
\end{tabular}
\label{tbl::app::key}
\end{table*}

To contextualize our estimation procedure, we acknowledge both its strengths and limitations. Our keyword-based counting method provides a transparent, scalable, and reproducible way to approximate the volume of research activity across domains, with higher keyword-mention thresholds generally yielding more reliable domain assignments. As shown in Table \ref{tab:threshold}, the number of papers decreases consistently as the threshold increases, reflecting increasing specificity but reduced recall. We report results primarily using the threshold of keyword matches greater than five, though counts under alternative thresholds are also available. However, this approach also inherits several limitations: low thresholds may inflate counts by including papers that mention a dataset only in passing, while high thresholds may exclude legitimate work that uses a benchmark but references it sparsely. Additionally, cross-domain papers contribute to multiple categories, and naming variations may lead to undercounting unless normalized. Despite these caveats, the method remains a practical heuristic for capturing broad research trends in benchmark-driven LLM evaluation.

\subsection{Detailed Application Paper Study}

\subsubsection{LLMs for General QA Tasks}
LLMs demonstrate strong performance in general question answering by enhancing reasoning capabilities, mitigating hallucinations, and effectively leveraging external knowledge. Current research emphasizes sequential sub-question reasoning, robust answer generation under uncertainty, and retrieval-augmented methods to improve information access. The following sections present representative approaches in each of these areas, with a focus on model-tuning strategies. For studies that do not involve model tuning, we refer readers to \citet{ke2025surveyfrontiersllmreasoning}.

\textbf{Step-by-Step Reasoning.}
Most works decompose a question into sub-questions and train LLMs to answer them sequentially to produce more accurate CoTs. For instance, 
\citet{qa_4} proposes generating sub-questions with GPT for each sentence in the ground-truth answer, while \citet{qa_9, qa_14} use model rollouts to create preference data based on the quality of the final answers. Furthermore, \citet{qa_11, qa_17} align LLM behavior in deciding when to reflect, answer, or ask follow-up questions. 

\textbf{Hallucination Management.}
Several works investigate how LLMs behave under uncertainty, particularly regarding hallucination. For instance, \citet{qa_2} treats GPT generations as positive samples for RLHF, whereas \citet{qa_6} suggests using conservative RL (assigning low scores when the model is unfamiliar) to mitigate hallucination during RLHF. Other approaches \citep{qa_1, qa_3, qa_7, qa_20} propose alternative data pipelines for the RL stage, yielding more robust answers to challenging follow-ups and enabling refusal in cases of unknown scenarios. Moreover, \citet{qa_8, qa_12} equip the LLM with confidence estimation capabilities, typically synthesized from multiple rollouts and GPT-generated root causes.

\textbf{LLM-based Retrieval.}
\citet{qa_10, qa_16, qa_18, qa_21, qa_15} post-train LLMs' query generation with retrieval-based rewards, leveraging them as query rewriters to improve retrieval effectiveness. Other studies \citep{qa_5, qa_13} propose alternative evaluation criteria for extracting key information from retrieved documents. Additionally, \citet{qa_19} converts long documents into hierarchical graphs and applies RL to the policy model for RAG using mode-seeking DPO algorithms.

\subsubsection{LLMs for Mathematical Tasks}
Recent efforts have enhanced LLMs for mathematical reasoning to support multi-step logic, symbolic manipulation, and rigorous verification.

\textbf{Math-Based Capabilities.}
\citet{math_data_1, math_data_3, math_data_2} fine-tuned LLMs on mathematics-specific corpora
to enhance mathematical reasoning and proof generation. Building on this, \citet{math_evol_data_1} applied the Evol-Instruct framework to math question–answer datasets. 
Similarly, \citet{math_evol_data_2, math_evol__1} improved Q\&A pairs via rephrasing, backward reasoning, and other data augmentation strategies. In parallel, \citet{math_evol__0} distilled reasoning abilities from proprietary models by collecting successful solution trajectories. Additionally, \citet{shao2024deepseekmath} used FastText-based filtering in Common Crawl and employed GRPO to further enhance problem-solving performance.


\textbf{Rollout Selection and Verification.}
\citet{math_verify} proposed training separate verifier models to assess the correctness of model-generated solutions, selecting answers with the highest verification scores. \citet{math_verify_4} fine-tuned LLMs on high-confidence data, while \citet{math_verify_6} filtered successful self-generated solutions for iterative training. \citet{math_verify_2} introduced 
MCTS to generate rollouts and annotate intermediate reasoning steps with probabilities of reaching correct answers. Similarly, \citet{math_verify_3} learned a value function from MCTS probabilities to guide exploration and DPO data selection, with \citet{math_verify_6} further refining DPO via a math-specific critic model. Finally, \citet{math_RL_step} fine-tuned models to select actions 
during mathematical reasoning.


 \begin{table*}
\centering
\caption{Paper counts under different keyword-mention thresholds across four domains.}
\label{tab:threshold}
\resizebox{\textwidth}{!}{
\begin{tabular}{c|rrr|rrr|rrr|rrr}
\hline
 & \multicolumn{3}{c|}{\textbf{QA}} 
 & \multicolumn{3}{c|}{\textbf{Math}} 
 & \multicolumn{3}{c|}{\textbf{Agentic}} 
 & \multicolumn{3}{c}{\textbf{Code}} \\
\textbf{Keyword count} 
 & \textbf{23} & \textbf{24} & \textbf{25} 
 & \textbf{23} & \textbf{24} & \textbf{25}
 & \textbf{23} & \textbf{24} & \textbf{25}
 & \textbf{23} & \textbf{24} & \textbf{25} \\
\hline
>0  & 771 & 1644 & 1308 & 12673 & 20405 & 13177 & 1310 & 1826 & 1373 & 334 & 1091 & 1302 \\
>1  & 580 & 1321 & 1088 & 3870 & 7143 & 4889 & 330 & 541 & 437 & 209 & 767 & 921 \\
>2  & 453 & 1022 & 858 & 1673 & 3295 & 2684 & 162 & 310 & 260 & 178 & 621 & 717 \\
>3  & 384 & 861 & 730 & 932 & 1957 & 1953 & 125 & 234 & 207 & 148 & 549 & 604 \\
>4  & 338 & 746 & 621 & 655 & 1391 & 1587 & 108 & 196 & 167 & 130 & 481 & 513 \\
\textbf{>5}  & \textbf{292} & \textbf{652} & \textbf{558} & \textbf{492} & \textbf{1098} & \textbf{1366} & \textbf{100} & \textbf{174} & \textbf{145} & \textbf{115} & \textbf{428} & \textbf{458} \\
>10 & 165 & 342 & 326 & 221 & 538 & 794 & 68 & 122 & 92 & 69 & 254 & 252 \\
>20 & 56 & 114 & 90 & 97 & 195 & 302 & 34 & 71 & 46 & 37 & 126 & 93 \\
\hline
\end{tabular}
}
\end{table*}

\subsubsection{LLMs for Agentic Tasks}
Agentic tasks challenge LLMs to act as autonomous decision-makers in dynamic environments, requiring continuous state perception, contextual integration, and multi-step planning. Success hinges on long-term reasoning, history management, and adaptive feedback, as each action influences future outcomes.

\textbf{Autonomous Task Execution.}
\citet{schick2023toolformer, qin2023toolllm, sft_mcts_3, sft_data_3, sft_data_4, sft_data_0, sft_data_6, sft_data__1} developed custom pipelines to behavior-clone full trajectories with open-weight models, supporting task-specific capabilities, whereas \citet{sft_data_1, sft_data_7} fine-tuned instruction-following models using actor–reflector outputs from proprietary APIs. To improve execution reliability, \citet{sft_data_5} used LLM judges to filter low-quality actions, \citet{sft_data_8} excluded erroneous steps during loss computation, and \citet{sft_data_9} added reflective annotations. Extending this, \citet{sft_alg_1} introduced knowledge-based self-learning with dynamic memory. In reinforcement learning, \citet{rl_data_2} curated preference pairs by synthesizing reasoning traces for ground-truth actions, while others generated preference data iteratively: ETO \citep{rl_data_0} used successful and failed trajectories for DPO, and \citet{rl_data_3} applied MCTS with corrections at the first error step.




\textbf{Self-Improvement and Feedback Integration.}
Some studies use sequential SFT followed by RL. \citet{both_seq_1} leveraged LLM API reflections, while \citet{both_seq_2} showed early RL is effective after initial SFT using a 70B teacher model. Hierarchical RL approaches assign rewards at the sentence or utterance level while updating policies at the token level \citep{rl_alg_1, rl_alg_6}, with DMPO \citep{rl_alg_3} adding step-wise discounting and \citet{rl_alg_0} using critic-based curriculum learning. OREO \citep{rl_alg_5} enables multi-step reasoning via maximum-entropy learning, complemented by IRL and PPO tuning \citep{rl_alg_4} and reward calibration \citep{rl_alg_2}. Some works jointly train SFT and RL losses with variants including masked or weighted objectives, outcome- and step-based DPO, and pseudocode planning \citep{both_sametime_1, both_sametime_2, both_sametime_0, rl_alg_7}.

\textbf{Planning and Long-Horizon Control.}
\citet{both_seq_3} proposed a hierarchical planning framework with a top-level planner fine-tuned on seed plans and optimized via DPO. \citet{both_seq_4} added Rejected Sampling Fine-Tuning (RFT) to further refine high-reward actions, while \citet{rl_data_4} adjusted exploration and prioritized critical steps for DPO. For test-time scaling, \citet{sft_mcts_2} integrated general and agentic data \citep{sft_mcts_3} with multi-path reasoning, and \citet{sft_mcts_1} combined policy fine-tuning with reward-model training, showing that beam search with explicit reward modeling improved performance. 



\subsubsection{LLMs for Code Generation Tasks}
LLMs have shown remarkable potential in code generation, editing, and task-specific reasoning, which are often improved by structured code data, executable feedback, or multi-agent frameworks for more robust and verifiable outputs.  


\textbf{Code Generation Capabilities.}
Recent studies have explored methods to improve code generation with LLMs. \citet{code_gen_1} employ self-instruction to synthesize training data, while \citet{code_gen_2} generate harder queries by modifying constraints of seed questions. \citet{code_gen_3} leverage open-source snippets to produce diverse and controllable instruction data. \citet{code_other_4} show that preceding CoT reasoning with code explanations enhances accuracy. More recently, \citet{code_gen_4} introduce a multi-agent framework in which an SFT model generates rollouts for DPO training.


\textbf{Code Editing Capabilities.} 
Code editing focuses on refining code that fails to meet its intended functionality, whether human-written or LLM-generated. \citet{code_edit_6} use human pull requests to guide GRPO-based specialization of LLMs. \citet{code_edit_4, code_edit_2} generate refinement rollouts evaluated against unit tests. \citet{code_edit_1} decompose code generation into verifiable subtasks for fine-grained alignment, while \citet{code_edit_7} propose a four-stage pipeline combining rejection sampling with rule-based RL rewards. \citet{code_edit_5} fine-tune an auxiliary LLM to produce verbal critiques that assist the primary model in correcting errors.


\textbf{Code Reuse for Other Tasks.}
Code can also be leveraged to enhance a variety of tasks by serving as a verifiable and interpretable reward signal. For instance,
\citet{code_edit_3} use code execution to assess the instruction-following ability of LLMs. 
\citet{code_edit_8} train an SLM to leverage LLMs via environment feedback by simulating workflows as executable Python code. 
\citet{code_other_1, code_other_ds, code_other_2} perform supervised instruction tuning on code-augmented math or data science tasks.

\subsection{Trend Analysis Details} \label{appendix:fig_dis}
This subsection covers a detailed analysis of the procedure-wise trend from studied papers, summarized in Figure \ref{fig:app}.

\textbf{Rapid Growth Across Domains.} Between 2023 and 2025, research activity accelerates sharply across all major task domains, though the magnitude of growth varies substantially. QA studies more than double from 292 in 2023 to 652 in 2024 (+123\%), with projections indicating continued expansion to 983 in 2025 (+118\%). Math-related research grows even faster in absolute scale, rising from 492 to 1,098 (+123\%) and then reaching 2,399 papers in 2025, which is a near 5× increase from 2023. Agent-focused work, while smaller in volume, shows steady expansion from 100 to 179 (+79\%) and further to 261 (+46\%).
The most dramatic surge occurs in code-related studies, which climb from 115 in 2023 to 428 in 2024 (+272\%) and nearly double again to 786 in 2025 (+84\%). This trajectory reflects the rapid maturation of code-centric benchmarks, tools, and evaluation pipelines, and highlights coding as one of the most rapidly diversifying application domains for LLMs.

\textbf{Convergence Toward Hybrid Training.} The landscape of training methodologies undergoes a marked consolidation toward hybrid approaches that combine SFT with RL or other post-training strategies. In 2023, SFT dominates at 73.3\%, with hybrid (“Both”) methods representing only 20.0\%. By 2024, hybrid training expands by 269\%, becoming the most common approach at 73.8\%, superseding pure SFT (19.1\%) and growing further to 70.6\% in 2025.
This shift illustrates a collective movement toward more sophisticated multi-stage training pipelines that exploit complementary strengths of SFT and RL. Meanwhile, RL-only methods, which starts initially 6.7\% of studies, gain modest traction, reaching 7.1\% by 2024 and 11.8\% by 2025, reflecting ongoing improvements in open-source RLHF frameworks and increased accessibility of preference data.

\textbf{Shift from Proprietary to Open Models.} A pronounced realignment in model choice accompanies these methodological changes. Between 2023 and 2025, reliance on proprietary API-based models declines sharply. The trend starts at 32.2\% in 2023 to 19.9\% in 2024, and down to 11.1\% by 2025, where researchers pivot toward open-weight models and standardized benchmarks. Open-weight usage more than doubles, growing from 12.2\% in 2023 to 17.5\% in 2024 and reaching 25.0\% in 2025.
Benchmarks remain the most commonly used resource category but also increase steadily (48.9\% to 61.1\% over three years), underscoring the field’s movement toward reproducible, transparent experimentation. In parallel, the use of human-curated or web-scraped datasets continues to shrink (6.7\% to 2.8\%), reflecting both improved benchmark coverage and growing concerns around legality, license compliance, and data provenance.
Overall, these trends indicate a strong and persistent shift toward open, standardized, and reproducible workflows as community norms solidify.

\section{Hardware Requirements for SFT/RL}

For researchers intending to perform practical experimentation with post-training alignment, consulting community-reported hardware guidance can be valuable. It is important to note, however, that GPU requirements vary substantially depending on model size, training approaches, e.g., full fine-tuning versus parameter-efficient methods such as LoRA \citep{hu2021loralowrankadaptationlarge} or QLoRA \citep{dettmers2023qloraefficientfinetuningquantized}, choice of optimizer, and software frameworks. Moreover, these requirements continue to evolve alongside advances in hardware and software tools.

For SFT, a commonly cited heuristic indicates that full fine-tuning of a transformer-based large language model requires approximately 16GB of VRAM per billion parameters when using half-precision (FP16). This high memory demand arises from the need to store gradients and optimizer states, and is substantially greater than the roughly 2GB per billion parameters typically needed for inference. Parameter-efficient approaches, such as LoRA and QLoRA, can reduce this requirement significantly, often into the range of 5–20GB on contemporary GPUs, depending on model size and precision settings\footnote{\url{https://modal.com/blog/how-much-vram-need-fine-tuning}}.

For RL training, frameworks such as \emph{verl} provide community-reported guidance on hardware requirements across different model scales. Practical experimentation indicates that RL generally demands substantially more memory than SFT due to additional overhead from rollout generation, policy gradients, and reward models. Heuristic estimates suggest that medium-sized models (1–3B parameters) can be trained on a single 80GB GPU using parameter-efficient methods like LoRA, while larger models (7–14B) typically require 2–4 such GPUs. Very large models (32B+) often necessitate 4–8 high-memory GPUs, with full-parameter training potentially exceeding 600GB of aggregate VRAM. Roughly, LoRA-based RL training consumes ~15–25GB VRAM per billion parameters, compared with ~16GB/B for full SFT fine-tuning, illustrating that RL generally incurs 1.5–3× higher memory demands per parameter. These numbers should be treated as approximate guidelines, as actual requirements depend on batch size, rollout length, optimizer choice, and memory-saving strategies such as gradient checkpointing or quantization\footnote{\url{https://verl.readthedocs.io/en/latest/perf/device_tuning.html}}.

It is essential to treat these hardware guidelines as approximate starting points rather than strict requirements. Actual resource needs will depend on factors including model architecture, task complexity, batch size, rollout length, and ongoing improvements in software and hardware technologies.

\section{More References} \label{appendix:more_ref}
This section provides additional references that extend beyond the text-only scope emphasized in Section~\ref{sec:application}. These works highlight broader developments in multimodal learning, retrieval-augmented methods, hallucination mitigation, and domain-specific datasets and capabilities. Together, they offer a more comprehensive view of the rapidly evolving ecosystem surrounding LLM reasoning and application domains.

One of the most practically oriented directions is multimodal LLM research, which integrates modalities such as speech and vision beyond pure text. This line of work serves as a natural extension to the text-only tasks surveyed earlier, offering insights into how additional sensory channels introduce new modeling challenges and opportunities. For instance, \citet{cui2025recentadvancesspeechlanguage} and \citet{yang2025largelanguagemodelsmeet} summarize methods for leveraging and aligning LLMs with speech-augmented multimodal data, enabling richer understanding and generation capabilities.

Beyond multimodality, another cluster of related literature addresses system-level improvements that refine how LLMs access information and maintain reliability. \citet{abootorabi2025askmodalitycomprehensivesurvey} studied advances in retrieval-augmented generation (RAG) across various modalities, providing a complementary perspective to reasoning-centric approaches. Meanwhile, \citet{li2025knowledgeboundarylargelanguage} investigates hallucination through the lens of knowledge boundaries, and \citet{zhu-etal-2025-trust} explores hallucination in conjunction with multimodal trustworthiness. In parallel, \citet{he-etal-2025-breaking} review methodologies for constructing SFT and RL datasets specifically tailored for reasoning, bridging data collection practices with model behavior.

Finally, several specialized studies focus on domain-specific datasets and complex task settings, which enrich the understanding of how LLM reasoning varies across fields. For example, \citet{yan2025surveymathematicalreasoningera} discusses mathematical reasoning datasets and benchmarks that support structured and verifiable inference. Complementing this, \citet{wei2025plangenllmsmodernsurveyllm} reviews LLM-based planning techniques applicable to both question answering and agentic tasks. Further extending the discussion to environments requiring perception–action loops, \citet{hu-etal-2025-os, nguyen-etal-2025-gui} examine multimodal–agent interactions, including GUI-based interfaces that illustrate how reasoning, perception, and action co-evolve. Together, these studies broaden the contextual understanding of LLM capabilities beyond the domains highlighted in the main text.

\section{Use of AI Assistants}
We employed LLM-based tools exclusively for language polishing during manuscript preparation and for assisting with full-text filtering of SFT/RL-related papers for human review in the Applications section (Section~\ref{sec:application}). All outputs generated by LLMs were carefully reviewed by the authors, who take full responsibility for any potential errors or hallucinations.

The prompt used with \texttt{gpt-oss-120b} to identify and classify SFT/RL-related papers is provided in the following verbatim. For all prompt requests to this model, we use FP4 precision, a context length of 128k, a temperature of $0.8$, and set top-k and top-p to $40$ and $0.95$, respectively.

Human readers subsequently verified all candidate papers to confirm their publication status and the relevance of their training paradigms (i.e., whether they involve SFT, RL, or both).

\begin{verbatim}
You are a research paper analyzer.
From the given scientific text, extract and classify the following:

1. The **proposed method** (ignore ablations).
2. The **comparison methods/baselines** the paper evaluates against.
3. The **datasets/benchmarks** used for evaluation.
4. The **training type** of the proposed method, exactly one of:
   - Reinforcement Learning (RL)
   - Supervised Fine-Tuning (SFT)
   - Both RL and SFT (SFT+RL)
   - Prompt Optimization
   - none-text Modality (vision, speech, multimodal, etc)
   - Other Methods
   - Survey (no new method, only summarization of others)

Output format:
<thought>
YOUR THOUGHT
</thought>

<answer>
{
  "method": "...",
  "baselines": ["..."],
  "benchmarks": ["..."],
  "training_type": one of ["SFT", "RL", "SFT+RL", "Prompt", "Modality", "Other", "Survey"]
}
</answer>
\end{verbatim}

\end{document}